\documentclass{article}

\usepackage{microtype}
\usepackage{graphicx}
\usepackage{subfigure}
\usepackage{booktabs} 

\usepackage{amsmath}
\usepackage{wrapfig}
\usepackage[linesnumbered,ruled,vlined]{algorithm2e}
\usepackage[T1]{fontenc}
\usepackage{amssymb}
\usepackage{amsthm}

\theoremstyle{definition}

\usepackage{dsfont}
\usepackage{mathrsfs}
\usepackage{bm}

\SetStartEndCondition{ }{}{}%
\SetKwProg{Fn}{def}{\string:}{}
\SetKwFunction{Range}{range}
\SetKw{KwTo}{in}
\SetKwFor{For}{for}{\string:}{}%
\SetKwIF{If}{ElseIf}{Else}{if}{:}{elif}{else:}{}%
\SetKwBlock{myIf}{if}{}
\SetKwFor{While}{while}{:}{}%
\AlgoDontDisplayBlockMarkers\SetAlgoNoEnd\SetAlgoNoLine%

\usepackage{hyperref}

\usepackage[accepted]{icml2021}

\icmltitlerunning{Backpropagated Neighborhood Aggregation for Accurate Training of Spiking Neural Networks}

\begin{document}

\twocolumn[
\icmltitle{Backpropagated Neighborhood Aggregation\\ for Accurate Training of Spiking Neural Networks}




\begin{icmlauthorlist}
\icmlauthor{Yukun Yang}{ucsb}
\icmlauthor{Wenrui Zhang}{ucsb}
\icmlauthor{Peng Li}{ucsb}

\end{icmlauthorlist}

\icmlaffiliation{ucsb}{Department of Electrical and Computer Engineering, University of California, 
Santa Barbara, CA 93106. Y. Yang is an incoming Ph.D. student to University of California, Santa Barbara}

\icmlcorrespondingauthor{Peng Li}{lip@ucsb.edu}

\icmlkeywords{SNN, BP, BPTT, training, CIFAR10, Machine Learning, ICML}

\vskip 0.3in
]



\printAffiliationsAndNotice{}  
\begin{abstract}
While backpropagation (BP) has been applied to spiking neural networks (SNNs) achieving encouraging results, a key challenge involved is to backpropagate a  continuous-valued loss over layers of spiking neurons exhibiting discontinuous all-or-none firing activities.
Existing methods deal with this difficulty by introducing compromises that come with their own limitations, leading to potential performance degradation. We propose a novel BP-like method, called neighborhood aggregation (NA), which computes accurate error gradients guiding weight updates that may lead to discontinuous modifications of firing activities. NA achieves this goal by aggregating finite differences of the loss over multiple perturbed membrane potential waveforms in the neighborhood of the present membrane potential of each neuron while utilizing a new membrane potential distance function. 
Our experiments show that the proposed NA algorithm delivers the state-of-the-art performance for SNN training on several datasets.
\end{abstract}

\section{Introduction}
\label{sec:intro}

Artificial neural networks (ANNs) have achieved great progress on various  tasks including computer vision~\cite{krizhevsky2017imagenet}, neural language processing~\cite{brown2020language}, reinforcement learning~\cite{berner2019dota, ye2020towards}, and  other big data applications~\cite{niu2020dual}. Compared with the more conventional non-spiking ANNs, simply referred to as ANNs in this paper, spiking neural networks (SNNs) can exploit efficient temporal codes, exhibit a greater level of biologically plausibility, and achieve significantly improved energy efficiency on neuromorphic hardware ~\cite{merolla2014million, furber2014spinnaker, davies2018loihi}. While  recent SNN research advances have led to improved performance ~\cite{zenke2018superspike, shrestha2018slayer, wu2018spatio, jin2018hybrid, zhang2020temporal, kim2020unifying} particularly for tasks that are well suited for  SNNs~\cite{blouw2019benchmarking, deng2020rethinking}, SNN training  is still a challenging task in general. 

Backpropagation (BP) has been widely used for training ANNs. Recent works have also demonstrated the great potential of BP in training SNNs~\cite{huh2017gradient, bellec2018long, zenke2018superspike, shrestha2018slayer,  wu2018spatio, jin2018hybrid, wu2019direct, zhang2020temporal, kim2020unifying, yang2020temporal}. 
Nevertheless,  BP based training is complicated by issues such as spatiotemproal dynamics and discontinuities of firing activities in SNNs. Among these, one key challenge is to deal with non-differentiability of firing activities. 

To address this difficulty, two main families of BP methods,  activation-based (or surrogate gradient) and timing-based methods, and their combination have been proposed~\cite{kim2020unifying}. The activation-based  methods approximate the non-differentiable spiking neural activation function by substituting it with a smoothed model~\cite{zenke2018superspike, shrestha2018slayer, wu2018spatio, wu2019direct}, allowing error backpropagation  in the manner of backpropagation through time (BPTT) or its variants. However, smoothing the non-differentiable activation function effectively converts one actual spike to multiple fictitious spikes and blurs timing information~\cite{zhang2020temporal}, making learning with precise timing difficult.

On the other hand, timing-based methods train SNNs that code information using spike timing~\cite{bohte2002error, mostafa2017supervised,  zhang2020temporal} and are able to 
accurately compute the gradient of the loss with respect to firing times. Early timing-based methods~\cite{bohte2002error, mostafa2017supervised} are limited to SNNs in which each neuron only fires once. The recent development in this family~\cite{zhang2020temporal} can handle more general SNNs where neurons are allowed to fire multiple times, achieving the state-of-the-art results on several datasets including CIFAR-10. 
The error gradient computed by the timing-based methods is based on a key assumption: the firing count of each neuron remains unchanged during the training process. In practice this does not prevent adding (removing) spikes to (from) the network during training as demonstrated in \cite{zhang2020temporal}, it may cause  degradation of weight update precision when the firing counts  vary. Moreover, additional tricks such as a warm-up process may be required to bring up the firing activity level of the network before a timing-based method can be applied~\cite{zhang2020temporal}. Ideas for combining the activation and timing-based methods have been suggested in \cite{kim2020unifying}.

We propose a new BP method called \emph{neighborhood aggregation} (NA) which acts as a general SNN training method while addressing the limitations of prior activation-based and timing-based methods. The all-or-none nature of spiking activities leads to the non-differentiability of the spiking activation function at firing threshold  $\vartheta$, which causes the following fundamental problem. Introducing  discrete jumps in the firing activities of an SNN by way of training is essential for achieving a given learning target, e.g., measured by the minimization of a spike-based loss function. However, the conventional error gradient computed by gradient-based training methods corresponds to an infinitesimal change in the loss with respect to the tunable parameters (weights). As such, it cannot accurately guide the introduction of essential discrete jumps in spiking activities nor reliability predict the impact of these jumps on the loss.   For activation-based BP methods, this problem surfaces when these methods attempt to compute the derivative of the activation function that does not always exist.
Furthermore, spiking models are commonly emulated  in discrete time with a finite time step size. Under this case, 
an infinitesimal change of the membrane potential $\boldsymbol{u}$ that does not move $\boldsymbol{u}$ across the firing threshold at any time, e.g., due to an  infinitesimal weight update during training, yields no change in the spike train of the neuron, hence having no impact on the training loss. This implies that: \textbf{1)} in order to reduce the training loss, the weight updates must be sufficiently large to alter the spiking activities of the network; and \textbf{2)} the surrogate gradient around $\vartheta$ computed in the activation-based methods is infinitesimal and does not properly characterize the desired dependency of the loss on the weights. 

More generally, we argue that error gradient
is not the \emph{best} quantity to look at for training spiking neural networks. 
As such, in the proposed neighborhood aggregation (NA) method, we resort to finite difference to compute what we refer to as the \emph{aggregated gradient} for characterizing the dependency of the loss on the spiking neural membrane potentials, which also circumvents the difficulty in differentiating the activation function at the firing threshold. The proposed NA is underpinned by three key concepts/components: \textbf{1)} the membrane potential neighborhood $\mathbf{N}_{\boldsymbol{u}}=\{\boldsymbol{u}_p|p=1,\dots,M\}$ of a targeted membrane potential waveform $\boldsymbol{u}$ and the construction of $\mathbf{N}_{\boldsymbol{u}}$, \textbf{2)} a membrane potential distance function (MP-dist), and \textbf{3)} a computationally-efficient  backpropagated neighborhood aggregation pipeline that outputs the desired aggregated gradient
that can be employed to update the weights. Conceptually, $\mathbf{N}_{\boldsymbol{u}}$ consists of membrane potential waveforms $\boldsymbol{u}_p$ that are close to $\boldsymbol{u}$ and correspond to different spike trains. Hence, $\mathbf{N}_{\boldsymbol{u}}$ represents likely changes in the membrane potential waveforms introduced by one-step of weight update during training. We compute the dependency of the loss around the present membrane potential waveform $\boldsymbol{u}$ of each spiking neuron within its neighborhood  $\mathbf{N}_{\boldsymbol{u}}$ by aggregating the finite differences of the loss between $\boldsymbol{u}_i$ (each member of $\mathbf{N}_{\boldsymbol{u}}$)  and $\boldsymbol{u}$. The finite difference operations rely on the proposed distance measure MP-dist. Finally, NA is facilitated by a computationally efficient  backpropagated neighborhood aggregation pipeline that leverages the proposed membrane potential neighborhood concept and MP-dist. This pipeline computes the desired aggregated gradient for weight updates in the manner of standard BP methods.

Compared with the prior activation-based methods~\cite{zenke2018superspike, shrestha2018slayer, wu2018spatio, wu2019direct}, the proposed NA method does not attempt to differentiate the non-differentiable spiking activation function; instead it computes the aggregated gradient that is more desirable for SNNs exhibiting all-or-none firing characteristics. As a result, it mitigates the poor timing precision of the activation-based methods~\cite{zhang2020temporal, kim2020unifying}.  One the other hand, NA does not rely on the assumption of the existing timing-based methods~\cite{mostafa2017supervised, zhang2020temporal} that the firing count of each spiking neuron remains constant during training.  Hence, the aggregated gradient computed by NA is able to guide weight updates to not only alter the timing of the spikes but also add/remove spikes to minimize the training loss. 

Benchmarked by commonly adopted datasets including MNIST~\cite{lecun1998mnist}, N-MNIST~\cite{orchard2015converting}, and CIFAR10~\cite{krizhevsky2009learning}, the proposed NA algorithm  has been shown to outperform the existing state-of-the-art activation and timing-based BP methods for direct training of spiking neural networks and achieve high-accuracy spike computation with low latency.

\section{Background}
\label{sec:background}
\begin{figure*}[t]
    \centering
    \includegraphics[width=\textwidth]{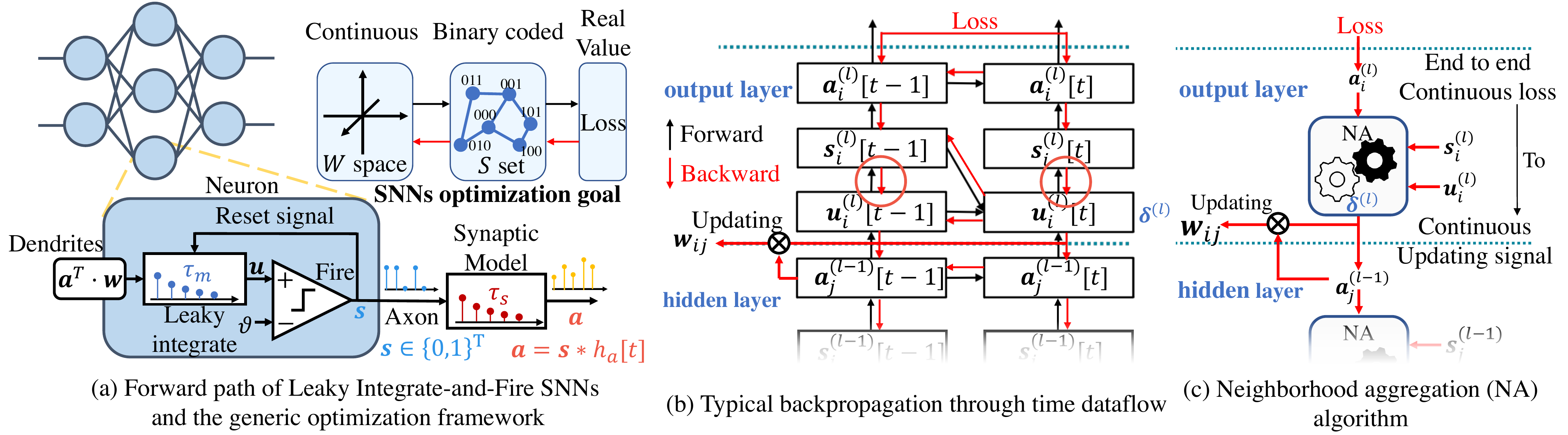}
    \vspace{-0.7cm}
    \caption{The forward model of leaky integrate-and-fire SNNs (a), and the backward mechanism comparison between (b) and (c).}
    \label{fig:na}
    \vspace{-0.5cm}
\end{figure*}
\subsection{The Spiking Neuron Model}

The leaky integrate-and-fire (LIF) neuron model~\cite{gerstner2002spiking}, one of the most prevalent choices for describing dynamics of spiking neurons, is adopted in this work. During simulation, we use the fixed-step first-order forward Euler method to discretize continuous membrane voltage updates over a set of discrete time steps.
The behavior of a discrete-time LIF neuron can be described by three variables: $\boldsymbol{a}$, the postsynaptic current (PSC), $\boldsymbol{u}$, the neuronal membrane potential, and $\boldsymbol{s}$, the output spike train. All neurons in an SNN share three parameters: $\tau_m$ -  membrane potential's  decaying time constant; $\tau_s$ - the time constant of the synaptic function; and $\vartheta$ - the firing threshold. The behavior of neuron $i$ in layer $l$ is described by:
\begin{equation}
\begin{aligned}
\boldsymbol{u}_i^{(l)}[t+1] &= \left(1-\frac{1}{\tau_m}\right)\boldsymbol{u}_i^{(l)}[t]\left(1-\boldsymbol{s}_i^{(l)}[t]\right)\\
&+\sum_{j=1}^{N^{(l-1)}} w_{ij}^{(l)}\boldsymbol{a}_j^{(l-1)}[t+1]
\label{eq:mem}
\end{aligned}
\end{equation}
\vspace{-0.2cm}
\begin{equation}
    \boldsymbol{a}_i^{(l)} = \boldsymbol{s}_i^{(l)} * \boldsymbol{\sigma},~~~ \boldsymbol{\sigma}[t]=\frac{1}{\tau_s}\left(1-\frac{1}{\tau_s}\right)^t
    \label{eq:a_conv}
\end{equation}
\begin{equation}
    \boldsymbol{a}_i^{(l)}[t+1]=\left(1-\frac{1}{\tau_s}\right)\boldsymbol{a}_i^{(l)}[t]+\left(\frac{1}{\tau_s}\right)\boldsymbol{s}_i^{(l)}[t+1]
    \label{eq:a_time}
\end{equation}
\begin{equation}
    \boldsymbol{s}_i^{(l)}[t] = H\left(\boldsymbol{u}_i^{(l)}[t]-\vartheta\right)
    \label{eq:spike}
\end{equation}
In (\ref{eq:mem}), the $1-\boldsymbol{s}_i^{(l)}[t]$ term reflects the effect of the firing-and-resetting mechanism and $w_{ij}^{(l)}$ is the weight of the synapse between neuron $j$ in layer $(l-1)$ and neuron $i$ in layer $l$.  The spike train of each presynaptic neuron is processed by a first-order synaptic kernel $\boldsymbol{\sigma}$ with  time constant ${\tau_s}$~\cite{gerstner2002spiking} to generate the post-synaptic current (PSC) to the postsynaptic neuron as shown in (\ref{eq:a_conv}), where (*) represents the time convolution. The same equation can also be written in the form of (\ref{eq:a_time}). (\ref{eq:spike}) defines the all-or-none firing activation function for which $H(\cdot)$ represents the Heaviside step function. 
The synaptic input integration and action potential generation process of an spiking neuron is shown at the bottom of Figure \ref{fig:na} (a), where the neuron is modeled as a nonlinear time-invariant system.

\subsection{The Loss Function for Supervised Training}
We use the Van Rossum distance~\cite{rossum2001novel} with the same kernel $\boldsymbol{\sigma}$ used in (\ref{eq:a_conv}) to measure the difference between the desired output spike-train $\boldsymbol{d}$ and the actual output spike-train $\boldsymbol{s}$ of each output neuron. We denote the loss function by $L = \sum_{t=0}^{N_t}E[t]$, where $N_t$ is the number of time steps and $E[t]$ is the loss at $t$. Specifically,   $L$ is given by:
\begin{equation}
\begin{aligned}
    L = \sum_{t=0}^{N_t}E[t] &= \sum_{t=0}^{N_t}\frac{1}{2}\left((\boldsymbol{\sigma}\ast \boldsymbol{d})[t]-(\boldsymbol{\sigma}\ast \boldsymbol{s})[t]\right)^2\\
    &=\sum_{t=0}^{N_t}\frac{1}{2}\left((\boldsymbol{\sigma}\ast \boldsymbol{d})[t]- \boldsymbol{a}[t]\right)^2
    \label{eq:Loss}\\
\end{aligned}
\end{equation}
Other differentiable loss functions are also usable. 

\subsection{Backpropagation Flow}
Figure \ref{fig:na} (b) shows the spatiotemporal dependencies which shall be considered while training an SNN using a method such as  backpropagation through time (BPTT)~\cite{shrestha2018slayer, zhang2020temporal}, where the loss is defined based upon the postsynaptic currents (PSCs) of the output neurons since the same synaptic function kernel is used to define the Van Rossum distance as mentioned before.

\vspace{-0.5cm}
\paragraph{Output layer.} Without loss of generality we differentiate the right hand of  (\ref{eq:Loss}), our chosen loss, with respect to each  $\boldsymbol{a}_i^{(l)}[t]$ and denote the resulting derivative by $\boldsymbol{g}_i^{(l)}[t]$: 
\begin{equation}
   \boldsymbol{g}_i^{(l)}[t]  = \boldsymbol{a}_i^{(l)}[t] - (\boldsymbol{\sigma}\ast \boldsymbol{d}_i)[t]
    \label{eq:output_error}
\end{equation}
Taking into account the temporal dependency in (\ref{eq:a_time}), we define
$\boldsymbol{e}_i^{(l)}[t]=\frac{\partial L}{\partial \boldsymbol{a}_i^{(l)}[t]}$,
which is called a PSC error signal:
\begin{equation}
    \label{eq:e}
    \boldsymbol{e}_i^{(l)}[t]=\left\{
    \begin{array}{lr}
    \boldsymbol{g}_i^{(l)}[t],& t=N_t \\
     \boldsymbol{g}_i^{(l)}[t]+\frac{\partial\boldsymbol{a}_i^{(l)}[t+1]}{\partial\boldsymbol{a}_i^{(l)}[t]}\boldsymbol{e}_i^{(l)}[t+1],& t<N_t
    \end{array}\right.
\end{equation}
where $\frac{\partial\boldsymbol{a}_i^{(l)}[t+1]}{\partial\boldsymbol{a}_i^{(l)}[t]} = \left( 1-\frac{1}{\tau_s} \right)$ according to (\ref{eq:a_time}).

We define $\boldsymbol{\delta}_i^{(l)}[t]= \frac{\partial L}{\partial \boldsymbol{u}_i^{(l)}[t]}$, denoting the backpropagated error for neuron $i$'s membrane potential at time $t$:
\begin{equation}
    \label{eq:delta}
    \boldsymbol{\delta}_i^{(l)}[t]  =\left\{
    \begin{array}{lr}
    \boldsymbol{e}_i^{(l)}[t]\frac{\partial \boldsymbol{a}_i^{(l)}[t]}{\partial \boldsymbol{u}_i^{(l)}[t]},& t=N_t \\
    \boldsymbol{e}_i^{(l)}[t]\frac{\partial \boldsymbol{a}_i^{(l)}[t]}{\partial \boldsymbol{u}_i^{(l)}[t]} 
    + \boldsymbol{\delta}_i^{(l)}[t+1]\frac{\partial \boldsymbol{u}_i^{(l)}[t+1]}{\partial \boldsymbol{u}_i^{(l)}[t]},& t<N_t
    \end{array}\right.
\end{equation}

\paragraph{Hidden layers.} If layer $(l-1)$ is a hidden layer,  $\boldsymbol{g}_i^{(l-1)}[t]$  is defined to be the error backpropagated from layer $l$'s membrane potentials to each $\boldsymbol{a}_i^{(l-1)}[t]$ without considering the temporal dependency between $\boldsymbol{a}_i^{(l-1)}[t]$ at different time points per (\ref{eq:mem}):
\begin{equation}
\begin{aligned}
    \boldsymbol{g}_i^{(l-1)}[t] &= \sum_{p=1}^{N^{(l)}} \left(\frac{\partial u_p^{(l)}[t]}{\partial \boldsymbol{a}_i^{(l-1)}[t]}\frac{\partial L}{\partial u_p^{(l)}[t]} \right) \\
    &= \sum_{p=1}^{N^{(l)}} w_{pi}^{(l)}\boldsymbol{\delta}_p^{(l)}[t],
    \label{eq:hidden_error}
\end{aligned}
\end{equation}
where $N^{(l)}$ is the number of neurons in layer $l$, and 
$\boldsymbol{\delta}_i^{(l)}[t]$ is obtained from (\ref{eq:delta}). Recursively applying (\ref{eq:e}), (\ref{eq:delta}) and (\ref{eq:hidden_error}) backpropagates the errors across all hidden layers. 

According to (\ref{eq:mem}), the derivative of loss $L$ with respect to the presynaptic weight $w_{ij}^{(l)}$ of neuron $i$ in layer $l$ shall consider the dependence of  $L$ on the neuron's membrane potential $\boldsymbol{u}_i^{(l)}[t]$ at all times: 
\begin{equation}
    \label{eq:d_l_d_w}
  	\frac{\partial L}{\partial w_{ij}^{(l)}} = \sum_{t=0}^{N_t} \frac{\partial \boldsymbol{u}_i^{(l)}[t]}{\partial w_{ij}^{(l)}} \boldsymbol{\delta}_i^{(l)}[t] = \sum_{t=0}^{N_t} \boldsymbol{a}_j^{(l-1)}[t]\boldsymbol{\delta}_i^{(l)}[t]
\end{equation}
\subsection{Key Challenges in Backpropagation}
In (\ref{eq:delta}), important derivatives to be computed are $\frac{\partial \boldsymbol{a}_i^{(l)}[t]}{\partial \boldsymbol{u}_i^{(l)}[t]}$, which is due to the effect of each neuron's membrane potential on its output PSC, and $\frac{\partial \boldsymbol{u}_i^{(l)}[t+1]}{\partial \boldsymbol{u}_i^{(l)}[t]}$, which reflects the temporal evolution of the membrane potential. As shown by the red circles in Figure. \ref{fig:na}, they all need to go through the non-differentiable path $\frac{\partial \boldsymbol{s}_i^{(l)}[t]}{\partial \boldsymbol{u}_i^{(l)}[t]}$, representing the key challenges in backpropagation. 

To address the above problems,~\cite{shrestha2018slayer} approximates the non-differentiable spiking activation function  by a probability density function of spike state change, and does not consider the temporal dependencies of a membrane potential across multiple time points.  
\cite{wu2018spatio} substitutes the  activation function by a continuous function during backpropagation. Furthermore,  this work sets $\frac{\partial \boldsymbol{u}_i^{(l)}[t+1]}{\partial \boldsymbol{u}_i^{(l)}[t]} = (1-\frac{1}{\tau_m})$, omitting the non-differentiable part of the temporal evolution of membrane potentials. Moreover, both of these methods effectively alter the underlying spiking neuron model and firing times. The more recent work of \cite{zhang2020temporal} instead computes the derivative of the PSC w.r.t. the membrane potential via spike timing to avoid the problem of non-differentiability, enabling learning targeted temporal sequences with high timing precision. However, it cannot directly handle increment and decrement of spike counts during training and requires a warm-up process prior to the application of the high-precision BP based training~\cite{zhang2020temporal}.

As shown in Figure \ref{fig:na}(c), the proposed NA method addresses the above problems by computing 
an aggregated gradient for weights updates to adjust both spike timing and  spike counts  to minimize the training loss. Specifically, NA computes the critical error signals $\boldsymbol{\delta}_i^{(l)}[t]$ more properly than (\ref{eq:delta}) to allow for weight updates per (\ref{eq:d_l_d_w}).

\section{Methods}
\label{sec:methods}

\subsection{Neighborhood Aggregation (NA): Basic Ideas}
The key motivations behind the proposed neighborhood aggregation (NA) method are as follows. 
The conventional error gradient computed by gradient-based training methods corresponds to an infinitesimal change in the loss with respect to the tunable parameters (weights). As such, it cannot accurately guide the introduction of essential discrete jumps in spiking activities nor reliability predict the impact of these jumps on the loss.
Furthermore, for SNNs emulated in discrete time,  an incremental weight change does not always produce a sufficiently large perturbation to spiking neurons' membrane potential waveforms $\boldsymbol{u} \in \mathbb{R}^{N_t}$ over $N_t$ time points to alter their output spike trains $\boldsymbol{s} \in \{0,1\}^{N_t}$. If so,  the weight update makes no difference to the training loss. On the other hand, a finite difference of the loss with respect to a sufficiently large perturbation on a neuron's membrane potential 
can be evaluated even with a non-differentiable spiking activation function. When performed properly and efficiently,  finite differences offer a better weight updating strategy for modifying firing activities to reduce the loss. 

With respect to the present membrane potential waveform $\boldsymbol{u}$ of a spiking neuron, consider a membrane potential waveform  $\boldsymbol{u}'$  close to $\boldsymbol{u}$  and but corresponds to a different spike train $\boldsymbol{s}'$. We call $\boldsymbol{u}'$ a \emph{neighbor} of $\boldsymbol{u}$. While it makes sense to examine the variation of loss $L$ between $\boldsymbol{u}$ and $\boldsymbol{u}'$, however, we do this with a key distinction from all previous BP methods:   we use a finite difference as opposed to a derivative of  $L$ w.r.t. the membrane potential:
\begin{equation}
    f_{d(\boldsymbol{u},\boldsymbol{u}')}L=\frac{L(\boldsymbol{u}')-L(\boldsymbol{u})}{\mathrm{d_{MP}}(\boldsymbol{u}, \boldsymbol{u}')},
    \label{eq:loss_overview}
\end{equation}
where $\mathrm{d_{MP}}(\cdot, \cdot)$ is the proposed membrane potential distance \emph{MP-dist} discussed later. 
The finite difference $ f_{d(\boldsymbol{u},\boldsymbol{u}')}L$ is evaluated over two membrane potentials across $N_t$ time points.  

There are three key components in the proposed NA algorithm. First, to compute the finite difference of (\ref{eq:loss_overview}) requires a proper distance function between a given pair of two membrane potential waveforms $\boldsymbol{u}$ and  $\boldsymbol{u}'$. For this, we propose the  membrane potential distance \emph{MP-dist} that assess the difference between $\boldsymbol{u}$ and  $\boldsymbol{u}'$ over $N_t$ time points while considering the temporal dependency of each membrane potential. 
Second, we compute $L(\boldsymbol{u}')-L(\boldsymbol{u})$ using a first-order approximation to boost the efficiency of the proposed NA pipeline. 
Last but not least, a meaningful evaluation of dependency of loss $L$ on the membrane potential via  (\ref{eq:loss_overview}) requires proper selection of neighbor $\boldsymbol{u}'$. To do this robustly, we aggregate $f_{d(\boldsymbol{u},\boldsymbol{u}')}L$ over a set of  membrane potentials $\mathbf{N}_{\boldsymbol{u}}=\{\boldsymbol{u}_p|p=1,\dots, N\}$ that are close to $\boldsymbol{u}$ and correspond to different spike trains. We call this set the \emph{membrane potential neighborhood} of $\boldsymbol{u}$ and discuss ways to construct it. For each $\boldsymbol{u}$, this aggregation leads to a vector  $\widetilde{\nabla}_{\boldsymbol{u}}L $ defined over $N_t$ time points and called the \emph{aggregated gradient}. 
Ultimately,  $\widetilde{\nabla}_{\boldsymbol{u}}L $ replaces the error vector $\boldsymbol{\delta}_i^{(l)}$ 
in (\ref{eq:d_l_d_w}) to perform layer-by-layer backprogagation.


\subsection{Membrane Potential Space}

\begin{figure}[t]
    \centering
    \includegraphics[width=0.43\textwidth]{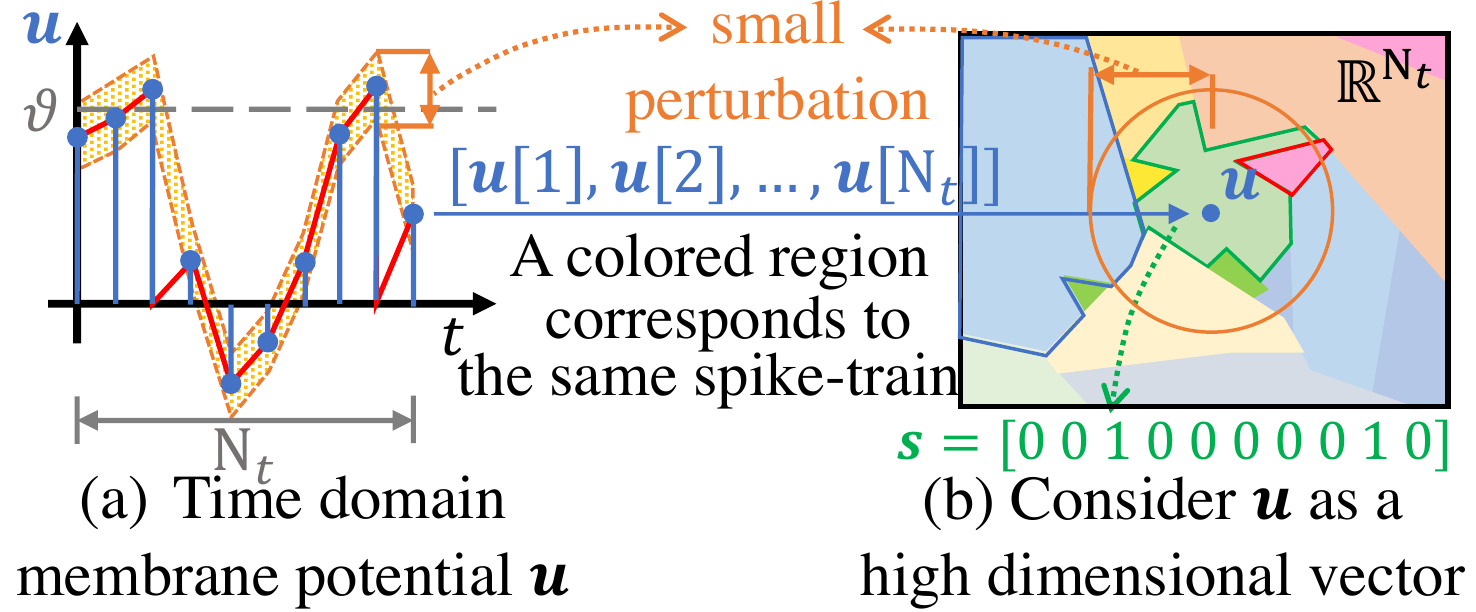}
    \vspace{-0.2cm}
    \caption{View membrane potential as a high dimensional vector.}
    \label{fig:method1}
    \vspace{-0.6cm}
\end{figure}

We consider the membrane potential $\boldsymbol{u}$ of a  neuron  as a high dimensional vector in  $\mathbb{R}^{N_t}$ as  in Figure \ref{fig:method1}. At each time point, whenever  $\boldsymbol{u}[t]$ exceeds the firing threshold $(\boldsymbol{u}[t]\geq\vartheta)$, a spike is generated. Conversely, whenever $\boldsymbol{u}[t]$ falls within the range of $\boldsymbol{u}[t]<\vartheta$, there is no spike. Therefore, membrane potential waveforms that correspond to the same spike train can be grouped together to  form a \emph{region} . All membrane potentials $\boldsymbol{u} \in \mathbb{R}^{N_t}$ are divided into many continuous regions corresponding to different spike trains represented by different colors in Figure \ref{fig:method1}.

\vspace{-0.5cm}
\paragraph{Key neighbors.} To compute the finite difference $f_{d(\boldsymbol{u},\boldsymbol{u}')}L$ of (\ref{eq:loss_overview}), we choose each $\boldsymbol{u}'$ belonging to a region $\boldsymbol{G}'$ different from region $\boldsymbol{G}$ where $\boldsymbol{u}$ is. This is to consider adding perturbations to $\boldsymbol{u}$  to create a change in the spike train.  While sharing the same spike train $\boldsymbol{s}'$, membrane potential waveforms  in  $\boldsymbol{G}'$ have a different distance to $\boldsymbol{u}$. 
We call  $\boldsymbol{u}_k' \in \boldsymbol{G}$ that is closest to $\boldsymbol{u}$  a \emph{key neighbor} of region $\boldsymbol{G}'$. Only key neighbors in different regions are chosen to form the membrane potential neighborhood of $\boldsymbol{u}$: $\mathbf{N}_{\boldsymbol{u}}=\{\boldsymbol{u}_p|p=1,\dots,n\}$. Practical ways for constructing $\mathbf{N}_{\boldsymbol{u}}$ is discussed in Section~\ref{sec:sns}.

\subsection{Membrane Potential Operations and Distance}\label{sec:MP-dist}
While distance concepts have been discussed for binary spike trains~\cite{spike_distances}, a distance function is needed to quantify the difference between a pair of two continuous-valued membrane potential waveforms $\boldsymbol{u}$ and $\boldsymbol{u}'$ over $N_t$ time points in  (\ref{eq:loss_overview}). Modifying the membrane potential $\boldsymbol{u}[t]$ of a neuron at time $t$ may impact $\boldsymbol{u}[t'], t'\geq t$ due to the dependencies introduced by the temporal evolution of  $\boldsymbol{u}$. To this end, the key drawback of applying existing distances for multiple-dimensional vectors such as the Hamming distance is that no temporal relationship 
among components of a vector is considered.



We propose a novel membrane potential distance \emph{MP-dist} denoted by $\mathrm{d_{MP}}(\cdot, \cdot)$ to consider temporal dependencies between the values of the membrane potential at different time points. According to the dynamics of LIF neurons in (\ref{eq:mem}),  membrane potential $\boldsymbol{u}_i^{(l)}$ of neuron $i$ can be obtained by integrating the total synaptic input $\boldsymbol{c}_i^{(l)} \in \mathbb{R}^{N_T}$, with $\boldsymbol{c}_i^{(l)}[t] = \sum_j w_{ij}^{(l)} \boldsymbol{a}_j^{(l-1)}[t]$, from a known initial condition. For the convenience of discussion, we define the function that maps from a given total synaptic input $\boldsymbol{c}$ to the corresponding membrane potential $\boldsymbol{u}$:  $f(\boldsymbol{c})=\boldsymbol{u}$. This mapping can be simply obtained by computing  $\boldsymbol{u}$ using (\ref{eq:mem}) to (\ref{eq:spike}). The inverse of $f$ maps from $\boldsymbol{u}$ to $\boldsymbol{c}$: $f^{-1}(\boldsymbol{u})=\boldsymbol{c}$. Note that $\boldsymbol{c}$ is readily available in each forward pass during training such that  $f^{-1}$ need not be explicitly constructed. 

\begin{algorithm}[b]
    \SetAlgoLined
     \textbf{Inputs:} $\boldsymbol{u}$, $\boldsymbol{\epsilon}$;~~~
     \textbf{Output:} $\boldsymbol{u}'$; ~~~ $\vartheta$: firing threshold;\\
    \textbf{Initialization:} $u=0$;\\
    $\boldsymbol{c}=f^{-1}(\boldsymbol{u})$; \\
    \For{$~t~{\rm in}~range(1, N_t)$}{
        $\boldsymbol{u}'[t] = u\left(1-\frac{1}{\tau_m}\right) + \boldsymbol{\epsilon}[t] + \boldsymbol{c} [t]$;\\
        $u = \boldsymbol{u}'[t] \left(1-H(\boldsymbol{u}'[t]- \vartheta)\right)$;
        }
    \Return{$\boldsymbol{u}'$}
    \caption{Membrane potential addition: $\boldsymbol{u}' =\boldsymbol{u}\boxplus \boldsymbol{\epsilon}$}
    \label{algo:add}
\end{algorithm}
\begin{algorithm}[t]
    \SetAlgoLined
    {Inputs:} $\boldsymbol{u}$, $\boldsymbol{u}'$;~~~
    {Output:} $\boldsymbol{\epsilon}$; ~~~ $\vartheta$: firing threshold; \\
    initialization: $u=0$ ;\\
    $\boldsymbol{c}=f^{-1}(\boldsymbol{u})$;\\
    \For{$~t~{\rm in}~range(1, N_t)$}{
        $\boldsymbol{u}[t]=u(1-\frac{1}{\tau_m}) + \boldsymbol{c} [t]$;\\
        $\boldsymbol{\epsilon} [t]=\boldsymbol{u}'[t]-\boldsymbol{u}[t]$;\\
        $u=\boldsymbol{u}'[t]\Big(1-H(\boldsymbol{u}' [t]-\vartheta)\Big)$;
    }
\Return{$\boldsymbol{\epsilon} $}
\caption{Membrane potential subtraction: $\boldsymbol{\epsilon} = \boldsymbol{u}'\boxminus \boldsymbol{u}$}
\label{algo:subtract}
\end{algorithm}

In the $N_t$-dimensional membrane potential space $\boldsymbol{U} =\mathbb{R}^{N_t}$, we define a commutative binary addition operator $\boxplus:\boldsymbol{U}\times\boldsymbol{U}\rightarrow \boldsymbol{U}$, which adds two given membrane potentials (vectors) $\boldsymbol{u}$ and $\boldsymbol{\epsilon}$ to produce a summed membrane potential $\boldsymbol{u}'  = \boldsymbol{u} \boxplus \boldsymbol{\epsilon}$ according to Algorithm~\ref{algo:add}. By making use of $\boxplus$, we can perturb the present membrane potential $\boldsymbol{u}$ by adding a perturbation $\boldsymbol{\epsilon}$ to it. Importantly, $\boxplus$ is not the element-wise addition. Instead,  $\boldsymbol{\epsilon}[t]$ is included in the total synaptic input to the neuron at time $t$ during the membrane potential integration process per (\ref{eq:mem}). Hence, $\boldsymbol{\epsilon}[t]$ may impact the resulting membrane potential at both the present and future times, i.e.,  $\boldsymbol{u}'[t'], t' \geq t$. In other words, $\boldsymbol{\epsilon}$ is added to $\boldsymbol{u}$ while considering the chain effects exerted on future time points. Under the same spirit, we define a binary subtraction operator $\boxminus:\boldsymbol{U}\times\boldsymbol{U}\rightarrow \boldsymbol{U}$ as described in Algorithm~\ref{algo:subtract}. $\boxminus$ finds the difference between a given pair of two membrane potentials $\boldsymbol{u}$ and $\boldsymbol{u}'$, which is $\boldsymbol{\epsilon} = \boldsymbol{u}' \boxminus \boldsymbol{u}$, such that $\boldsymbol{u}' = \boldsymbol{u}\boxplus (\boldsymbol{u}' \boxminus \boldsymbol{u} )$. Essentially, $\boxminus $ subtracts $\boldsymbol{u}$ from $\boldsymbol{u}'$ while considering the chain effects on future time points. It can be shown that $\boldsymbol{u}' \boxminus \boldsymbol{u} = - (\boldsymbol{u} \boxminus \boldsymbol{u}')$.

The proposed membrane potential distance $\mathrm{d_{MP}}(\cdot, \cdot):\boldsymbol{U}\times\boldsymbol{U}\rightarrow [0,+\infty)$ is a metric in the membrane potential space $\boldsymbol{U}=\mathbb{R}^{N_t}$ and is defined by $\mathrm{d_{MP}}(\boldsymbol{u}, \boldsymbol{u}') = ||\boldsymbol{u}' \boxminus \boldsymbol{u}||_2$.  Unitizing $\boxplus$, $\boxminus$, and  $\mathrm{d_{MP}}(\cdot, \cdot)$, one can find  key neighbors of the present membrane potential $\boldsymbol{u}$ by adding perturbations to it, form  the membrane potential neighborhood  $\mathbf{N}_{\boldsymbol{u}}$ of $\boldsymbol{u}$, and determine the distances between $\boldsymbol{u}$ and the members of $\mathbf{N}_{\boldsymbol{u}}$ for computing the finite differences of Loss $L$.

\subsection{Simple Neighborhood Selection (SNS)}\label{sec:sns}
\begin{figure}[b!]
    \vspace{-0.5cm}
    \centering
    \includegraphics[width=0.43\textwidth]{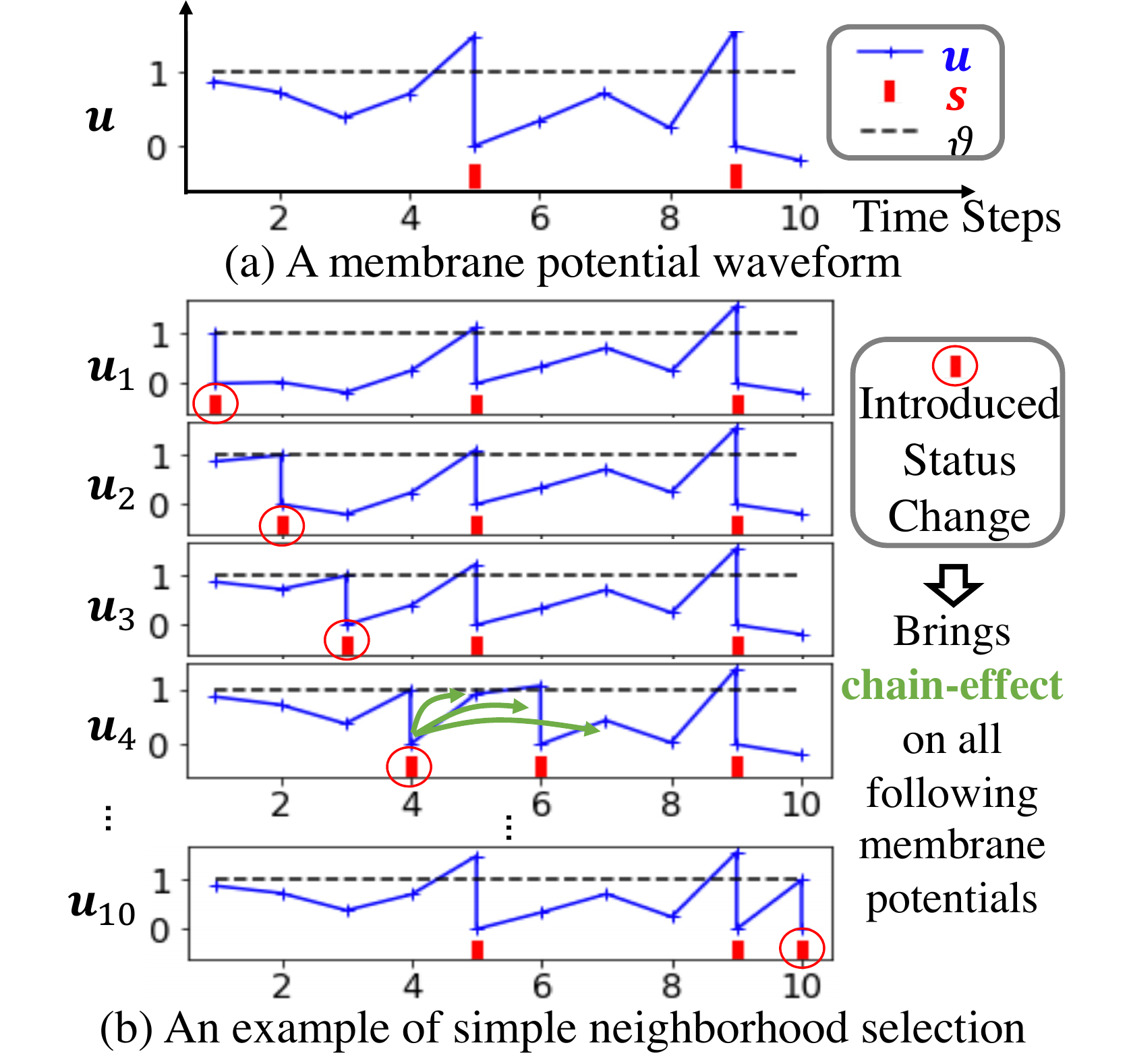}
    \vspace{-0.3cm}
    \caption{Simple neighborhood selection (SNS).}
    \label{fig:SNS}
\end{figure}
To well represent the landscape of the loss in the membrane potential space, for each spiking neuron one may choose $n$ closest neighbors evaluated by $\mathrm{d_{MP}}(\cdot, \cdot)$ to form a neighborhood $\mathbf{N}_{\boldsymbol{u}}$ of the present membrane potential $\boldsymbol{u}$.    However, it is not obvious how to design an algorithm to find the top $n$ nearest key neighbors from a total of $2^{N_t}$ regions in the membrane potential space with an acceptable computational complexity. 
As a first step, we propose a simple and effective strategy named SNS (Simple Neighborhood Selection). To select neighbors that are close to $\boldsymbol{u} \in \mathbb{R}^{N_t}$, SNS introduces a set of $N_t$ membrane potential perturbations:  $\boldsymbol{\epsilon}_{(i)}, i = 1, \cdots, N_t$, where $\boldsymbol{\epsilon}_{(i)}[j] = 0, \forall j \neq i$ and  $\boldsymbol{\epsilon}_{(i)}[i] \neq 0$. With $\boldsymbol{\epsilon}_{(i)}$ only a non-zero perturbation is introduced at time point $i$.  $\boldsymbol{\epsilon}_{(i)}[i]$ is chosen to be the least amount of membrane potential perturbation that can cause a firing status change, i.e., firing to no firing or vice versa, at time point $i$.  The initialized firing status change trickles down to later time points, causing additional membrane potential changes.  This chain-effect can be captured by Algorithm~\ref{algo:add}, which effectively produces a perturbed membrane potential  $\boldsymbol{u}_{(i)} = \boldsymbol{u} \boxplus \boldsymbol{\epsilon}_{(i)}$. We construct a $N_t$-member neighborhood $\mathbf{N}_{\boldsymbol{u}}=\{\boldsymbol{u}_{(i)}|i=1,\dots,N_t\}$. This process is illustrated in Figure \ref{fig:SNS} (b). The exploration of other neighborhood selection techniques is left for future work. 



\subsection{The NA Algorithm Pipeline}
Processing of a single network layer $l$ by the NA algorithm consists of five steps. The time argument and neuron and layer indices are dropped to simplify notations when causing no confusion. 



\textbf{Step 1. PSC Error Signal Computation}:
(\ref{eq:output_error}), (\ref{eq:e}), and (\ref{eq:hidden_error}) are on the differentiable path. 
Combining (\ref{eq:output_error}) and (\ref{eq:hidden_error}) gives:
\begin{equation}
\label{eq:g}
    \boldsymbol{g}_i^{(l)} = \left\{
    \begin{array}{lr}
    \boldsymbol{a}_i^{(l)} - (\boldsymbol{\sigma}\ast \boldsymbol{d}_i), &{\rm output~layer}\\
    \sum_{p=1}^{N^{(l+1)}}w_{ji}^{(l+1)}\boldsymbol{\delta}_p^{(l+1)}, &{\rm hidden~layer}
    \end{array}
    \right.
\end{equation}
Plugging (\ref{eq:g}) into (\ref{eq:e}) produces the desired PSC error signal $\boldsymbol{e}_i^{(l)}$ for neuron $i$, which will used in Step 4.

\textbf{Step 2. Neighborhood Selection}:
Following our earlier discussions, for each neuron one may choose $M$ key neighbors of the present membrane potential, corresponding to different spike trains.
This can be done by adding perturbations to the  membrane potential using SNS.

\textbf{Step 3. Membrane Potential Distance Computation}:
The distance between the present membrane potential $\boldsymbol{u}$ of a neuron and each key neighbor $\boldsymbol{u}_p$ in its neighborhood $\mathbf{N}_{\boldsymbol{u}}$ is determined using $\mathrm{d_{MP}}(\cdot, \cdot)$  through Algorithm \ref{algo:subtract}.  

\textbf{Step 4. First-order Loss Approximation}: Per (\ref{eq:loss_overview}), it is necessary to evaluate the loss
$L(\boldsymbol{u}_p)$ at each neighbor $\boldsymbol{u}_p$ to compute the variation of the loss, which can be done by performing
an additional forward propagation from the current layer to the output layer. Such repeatedly forward propagations for all neurons' neighborhoods add an unacceptable computational overhead to the training process. 
Noting that $L(\boldsymbol{u}_p) = L(\boldsymbol{a}(\boldsymbol{u}_p)) \equiv L(\boldsymbol{a}_p)$, we address this difficulty by 
using the PSC error signal $\boldsymbol{e} = \nabla_{\boldsymbol{a}} L$ given by (\ref{eq:e}) to estimate the variation of the loss due to the perturbation introduced to the PSC $\boldsymbol{a}$.  It is important to note that $\boldsymbol{e}$ is readily available in \textbf{Step 1}.

\begin{figure}[t]
    \centering
    \includegraphics[width=0.4\textwidth]{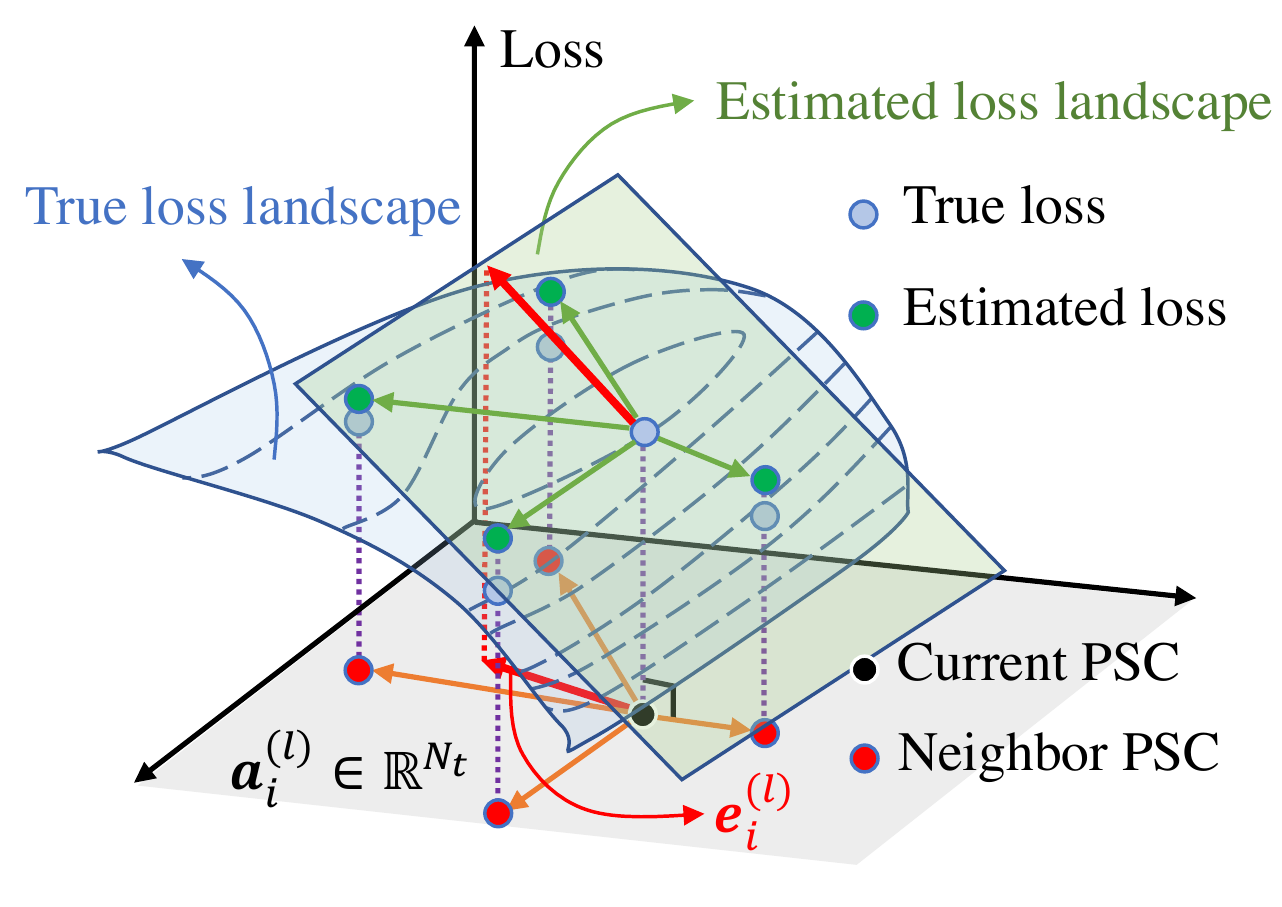}
    \vspace{-0.3cm}
    \caption{First order loss approximation}
    \label{fig:method4}
    \vspace{-0.4cm}
\end{figure}

As shown in Figure \ref{fig:method4}, using $\boldsymbol{e}\in\mathbb{R}^{N_t}$ for a first-order approximation  means approximating the loss landscape by a tangent hyper-plane around the present PSC $\boldsymbol{a}$. 
The first-order approximation of the change of the loss  between the current PSC $\boldsymbol{a} $ and one of its near neighbor's PSC $\boldsymbol{a} _p$ is:
\begin{equation}\label{eqn_first_order_approx}
\begin{aligned}
    L\left(\boldsymbol{a} _p\right) - L\left(\boldsymbol{a} \right) 
    & \approx \nabla_{\boldsymbol{a}} L\left(\boldsymbol{a} _p-\boldsymbol{a} \right)\\
    & = \boldsymbol{e}\cdot \left(\boldsymbol{a} _p-\boldsymbol{a} \right)
\end{aligned}
\end{equation}
As such, NA achieves great computational efficiency via first-order approximations of loss changes without expensive forward propagations.

\textbf{Step 5. Neighborhood Aggregation and Weight Updates}:
As discussed earlier in terms of (\ref{eq:loss_overview}), instead of attempting to compute the conventional gradient ${\nabla}_{\boldsymbol{u}} L \in \mathbb{R}^{N_t}$  which does not exist everywhere for spiking neurons, we compute 
the finite difference $ f_{d(\boldsymbol{u},\boldsymbol{u}_p)}L $ between $\boldsymbol{u}$ and one of its near neighbor $\boldsymbol{u}_p$ while using (\ref{eqn_first_order_approx}):
\begin{equation}
    f_{d(\boldsymbol{u},\boldsymbol{u}_p)}L=\frac{L(\boldsymbol{u}_p)-L(\boldsymbol{u})}{\mathrm{d_{MP}}(\boldsymbol{u}, \boldsymbol{u}_p)} \approx \frac{\boldsymbol{e}\cdot \left(\boldsymbol{a} _p-\boldsymbol{a} \right)}{\mathrm{d_{MP}}(\boldsymbol{u}, \boldsymbol{u}_p)},
    \label{eqn_finite_diff}
\end{equation}
which only captures the variation of loss between $\boldsymbol{u}$ and $\boldsymbol{u}_p$. 
Here we compute an aggregated gradient $\widetilde{\nabla}_{\boldsymbol{u}} L \in \mathbb{R}^{N_t}$ over the neuron's entire $M$-member neighborhood based on all   $f_{d(\boldsymbol{u},\boldsymbol{u}_p)}L$, where $p=1,2,\dots,M$. For this, we consider each  $f_{d(\boldsymbol{u},\boldsymbol{u}_p)}L$ as the directional derivative resulted from projecting $\widetilde{\nabla}_{\boldsymbol{u}} L$ onto the unit perturbation vector $\boldsymbol{o}_p = \frac{\boldsymbol{u}_p \boxminus \boldsymbol{u}} {||\boldsymbol{u}_p \boxminus \boldsymbol{u}||_2} =  \frac{\boldsymbol{u}_p \boxminus \boldsymbol{u}} {\mathrm{d_{MP}}(\boldsymbol{u}, \boldsymbol{u}_p)}$, leading to a system of linear equations on the left of (\ref{eq:pseudoiniverse}):
\begin{small}
\begin{equation}
    \begin{bmatrix} {\boldsymbol{o}}_1^T \\ {\boldsymbol{o}}_2^T \\\vdots\\ {\boldsymbol{o}}_M^T \end{bmatrix}\cdot\widetilde{\nabla}_{\boldsymbol{u}} L=\begin{bmatrix} f_{d(\boldsymbol{u},\boldsymbol{u}_1)}L\\f_{d(\boldsymbol{u},\boldsymbol{u}_2)}L\\\vdots\\
    f_{d(\boldsymbol{u},\boldsymbol{u}_M)}L\end{bmatrix}, \widetilde{\nabla}_{\boldsymbol{u}} L=\begin{bmatrix} {\boldsymbol{o}}_1^T \\ {\boldsymbol{o}}_2^T \\\vdots\\ {\boldsymbol{o}}_M^T \end{bmatrix}^+
    \begin{bmatrix} f_{d(\boldsymbol{u},\boldsymbol{u}_1)}L\\f_{d(\boldsymbol{u},\boldsymbol{u}_2)}L\\\vdots\\
    f_{d(\boldsymbol{u},\boldsymbol{u}_M)}L\end{bmatrix}
    \label{eq:pseudoiniverse}
\end{equation}
\end{small}
 $\widetilde{\nabla}_{\boldsymbol{u}} L$ is solved according to the right side of  (\ref{eq:pseudoiniverse}).
$+$ is the Moore–Penrose pseudoinverse considering that the system matrix may not always be invertible.

Using $\widetilde{\nabla}_{\boldsymbol{u}} L$ to substitute $\boldsymbol{\delta}$ in (\ref{eq:d_l_d_w}) allows for weight updates of layer $l$, after which  the backprogagation proceeds to the  preceding layer.


\subsubsection{Further simplifications under SNS}
The simplicity of neighborhood selection using SNS brings additional advantages. First, the MP-dist between $\boldsymbol{u}$ and any of its neighbors $\boldsymbol{u}_p$ in the formed neighborhood is simply: $\mathrm{d_{MP}}(\boldsymbol{u},\boldsymbol{u}_p)= \vartheta-\boldsymbol{u}[p]$. 
The $M$ unit perturbation vectors $\{{\boldsymbol{o}}_1,{\boldsymbol{o}}_2,\dots,{\boldsymbol{o}}_M\}$ in (\ref{eq:pseudoiniverse}), where $M=N_t$, form an identity matrix. As a result, the right side of (\ref{eq:pseudoiniverse})  reduces to:
\begin{equation}
    \widetilde{\nabla}_{\boldsymbol{u}} L=
    \begin{bmatrix} f_{d(\boldsymbol{u},\boldsymbol{u}_1)}L\\f_{d(\boldsymbol{u},\boldsymbol{u}_2)}L\\\vdots\\
    f_{d(\boldsymbol{u},\boldsymbol{u}_M)}L\end{bmatrix}
    \label{eq:summing_up}
\end{equation}
Hence, the  aggregated gradient $  \widetilde{\nabla}_{\boldsymbol{u}} L$ can be efficiently obtained without computing the pseudoinverse in the right side of (\ref{eq:pseudoiniverse}).

\section{Experimental Results}
We designed several experiments to demonstrate the effectiveness of the proposed membrane potential distance MP-dist and the NA algorithm while comparing with several other SNN training algorithms. 
Moreover, we provided a time-complexity analysis of several algorithms  with a wall-clock time comparison.

\subsection{MP-dist Quality Evaluation}

The finite difference based aggregated gradient computation highly relies on an accurate measure of distance between a given membrane potential $\boldsymbol{u}$ and its neighbor $\boldsymbol{u}_p$. While inducing discrete jumps in firing activity, small weight updates in training shall modify the present $\boldsymbol{u}$ to a close neighbor $\boldsymbol{u}_p$ that corresponds to a small MP-dist from $\boldsymbol{u}$.

\begin{figure}[b]
    \centering
    \vspace{-0.4cm}
    \includegraphics[width=0.48\textwidth]{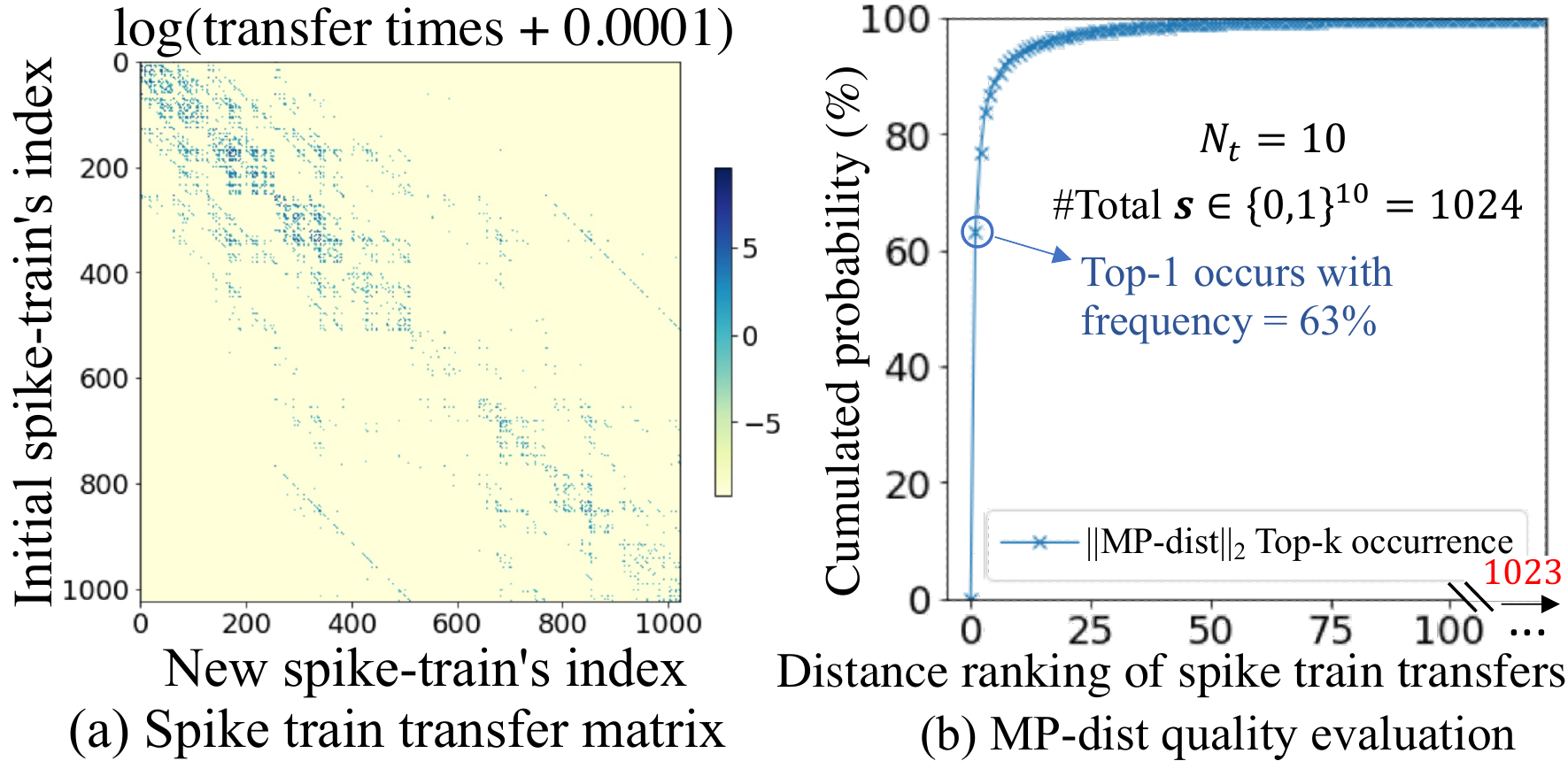}
    \vspace{-0.6cm}
    \caption{The quality evaluation of MP-dist $\mathrm{d_{MP}}(\cdot, \cdot)$.}
    \label{fig:compare}
\end{figure}

To evaluate the quality of the MP-dist distance $\mathrm{d_{MP}}(\cdot, \cdot)$, we counted the changes of all neurons' spike-trains of the SNN with the same architecture used in Table \ref{tab:MNIST}. The SNN was trained by TSSL-BP\cite{zhang2020temporal} on the MNIST dataset \cite{lecun1998mnist} in 100 training iterations with a total number of time-steps $N_T=10$, as shown in Figure \ref{fig:compare} (a). The input samples were kept the same before and after an training iteration to count the spike train changes. The decimal index of a spike train was converted from its corresponding binary code (e.g. 
$[1, 1, 1, 1, 1, 1, 1, 1, 1, 1]\rightarrow 1023$, $
[0, 0, 0, 0, 0, 0, 0, 1, 0, 1]\rightarrow 5$).

Each pixel in Figure \ref{fig:compare} (a) represents the count of transfers from a particular spike train (y-index) before a training weight update iteration to another spike train (x-index) after the iteration. We accumulated the counts for all neurons' output spike trains in the 100 iterations. For example, ``$(4,5)=12$'' means that the transition from the output spike train  $[0, 0, 0, 0, 0, 0, 0, 1, 0, 0] (\rightarrow 4)$ to the output spike train $[0, 0, 0, 0, 0, 0, 0, 1, 0, 1] (\rightarrow 5)$ happened $12$ times.

The values on the main diagonal of the transfer matrix represent the total number of cases that the corresponding output spike train remained unchanged before and after an training iteration, which are very large and uninformative. For ease of visualization, we set these values to zero and also showed the transfer counts in log scale to make small values visible.


Figure~\ref{fig:compare} (b) shows the cumulative probability distribution (CDF) of the ranking of the distance between the spike train before a training iteration and the one after the iteration evaluated using $\mathrm{d_{MP}}(\cdot, \cdot)$ based on the data tracked over all possible discrete spike train transfers of all neurons. To more meaningfully visualize the CDF, we ranked the collected $\mathrm{d_{MP}}(\cdot, \cdot)$, and displayed the ranking in the ascending order along the x-axis of Figure~\ref{fig:compare} (b). 
Based on MP-dist, the spike train transfers with the least  $\mathrm{d_{MP}}(\cdot, \cdot)$ value represent transfers between the closet spike trains or within the top-1 nearest regions in the membrane potential space, which  constituted a high percentage ($63\%$) of all transfers observed. The CDF reaches to  $100\%$ quickly. This results confirm the quality of MP-dist: a large majority of the neurons' output spike trains indeed changed to a spike train with a small MP-dist value.

\subsection{A Single-neuron Network}

\begin{figure}[t]
    \centering
    \includegraphics[width=0.48\textwidth]{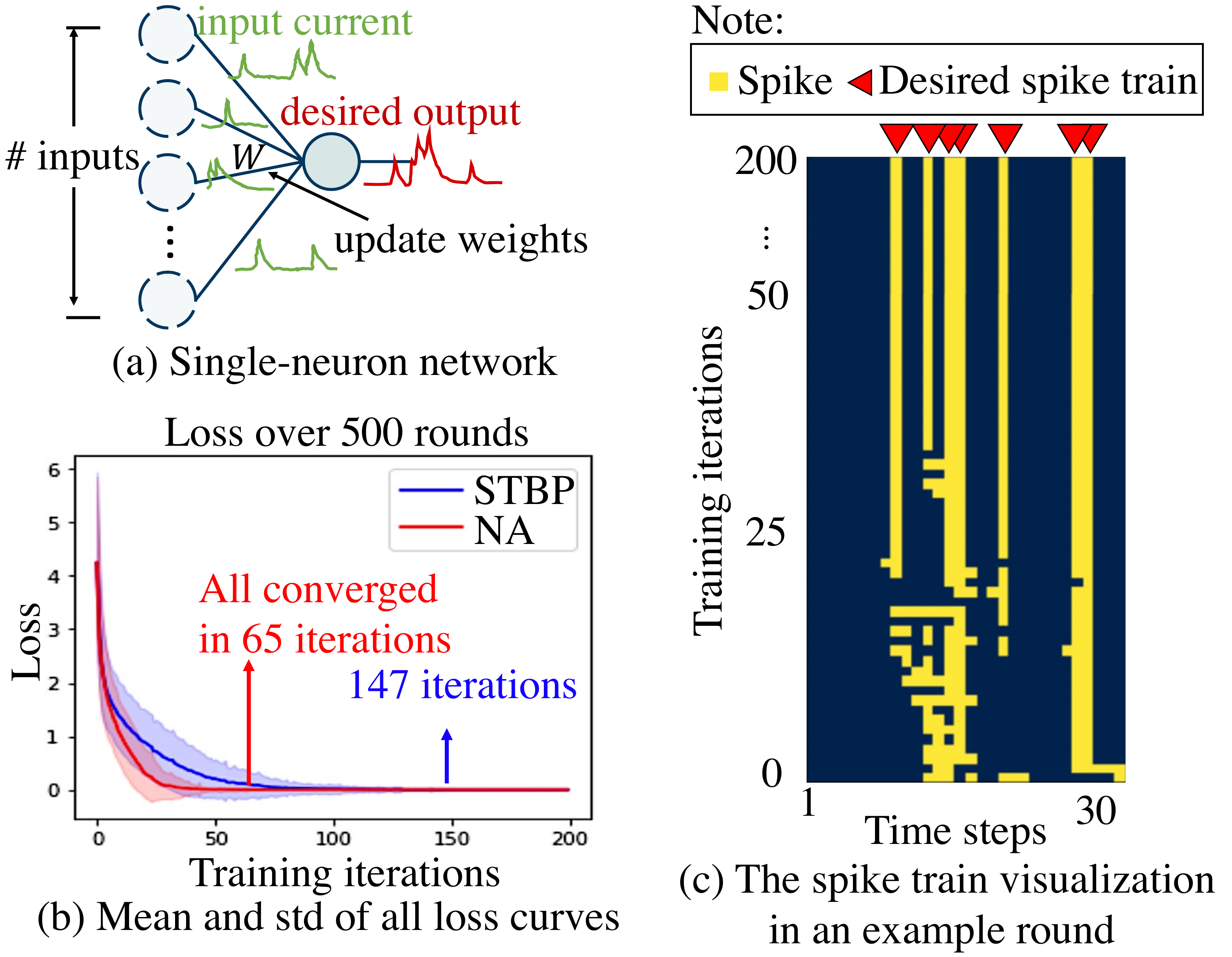}
    \vspace{-0.5cm}
    \caption{Training of the single-neuron network.}
    \label{fig:single_neuron}
\vspace{-0.6cm}
\end{figure}

This subsection demonstrates how the output neuron's spike-train changes towards the desired one in a single-neuron network (Figure \ref{fig:single_neuron} (a)) with two different learning algorithms.   
The parameters used are summarized in Table \ref{tab:single_param}. 500 independent training experiments (rounds), each with a different set of randomly generated spike inputs and desired output spike train, were conducted.  
\begin{table}[h]
    \centering
    \vspace{-0.5cm}
    \caption{Parameters of the single-neuron network.}    
    \resizebox{0.48\textwidth}{5mm}{
    \begin{tabular}{c c c c c c c c}
         \hline
         $N_T$  & \#inputs & $p_{in}$  &\#iters & \#rounds & $p_o$ &$\tau_s$ &$\tau_m$\\
         $30$  & $200$ & 0.05 & 200 & 500 & 0.2 & 2 & 5\\
         \hline
    \end{tabular}}
    \label{tab:single_param}
\end{table}
\vspace{-0.4cm}

In each round, we randomly generated spikes for all $200$ input neurons. The activity of the $i$-th input spike train at time step $t$, $\boldsymbol{s}_i[t]$, followed the Bernoulli distribution with $p(\boldsymbol{s}_i[t]=1)=p_{in}$,  for $i=1,2,...,200$. The PSC $\boldsymbol{a}_i$ generated by the $i$-th input neuron was computed based on a chosen synaptic function kernel $\boldsymbol{\sigma}$ by $\boldsymbol{a}_i=\boldsymbol{s}_i*\boldsymbol{\sigma}$, we use $\boldsymbol{\sigma}$ as in (\ref{eq:a_conv}), with $\tau_s=2$.  We then normalized 
using the mean and standard deviation of all PSCs. The input currents are then feed into the single output neuron with its membrane potential time constant $\tau_m=5$. Similarly, the desired output spike train $\boldsymbol{s}_o$ of the single output neuron, i.e., the label, was randomly generated following the Bernoulli distribution with probability $p(\boldsymbol{s}_o[t]=1)=p_o$.  Figure \ref{fig:single_neuron} (c) shows an example of how the neuron's output spike train changes towards the desired one with the NA algorithm in a single round.

We compared our NA algorithm with the recent activation-based surrogate gradient method STBP \cite{wu2018spatio,wu2019direct} by training the weights between the input neurons and the single output neuron for 200 iterations in each round. The loss function we use here is (\ref{eq:Loss}). We recorded the variation of the loss during the 200 iterations, and repeated the same experiment for 500 rounds. The recorded 500 loss curves are shown in Figure \ref{fig:single_neuron} (b). The solid lines are the mean values of the 500 loss curves  while the shaded areas illustrate the standard deviations. Clearly, the proposed NA algorithm reduced the losses quickly and could converge to the desired output firing patterns within 65 iterations. In contrast, STBP \cite{wu2019direct} did not all converge to the desired output patterns until 147 iterations. 

\subsection{The Results on the MNIST Dataset}
\begin{table}[b]
    \centering
    \vspace{-5mm}
    \caption{Performances of different SNN BP methods on MNIST.}    
    \begin{tabular}{c|c|c}
         \hline
         Method  & \#Steps & BestAcc \\
         \hline
         HM2BP~\cite{jin2018hybrid}  & 400 & 99.49\% \\
         ST-RSBP~\cite{zhang2019spike} & 400 & 99.62\% \\
         SLAYER~\cite{shrestha2018slayer}  & 300 & 99.41\% \\
         STBP~\cite{wu2018spatio} & 30 & 99.42\% \\
         TSSL-BP~\cite{zhang2020temporal}   & 5 & 99.53\% \\
         \textbf{This work} & \textbf{5} & \textbf{99.69\%} \\
         \hline
         \multicolumn{3}{l}{Spiking CNN structure: 15C5-P2-40C5-P2-300}
    \end{tabular}
    \label{tab:MNIST}
\end{table}
The proposed NA algorithm is compared with with several other SNN BP training methods for training a spiking convolutional neural network (CNN)  on the MNIST  dataset \cite{lecun1998mnist} in Table~\ref{tab:MNIST}. The NA algorithm used only a total of five time steps, same as for  TSSL-BP \cite{zhang2020temporal}, and gained the best accuracy among all these methods, showing its potential in accurate training of SNNs with short temporal latency.




\subsection{The Results on the N-MNIST Dataset}
The N-MNIST~\cite{orchard2015converting} is the neuromorphic version of the MNIST dataset~\cite{lecun1998mnist}. We adopted the same spiking CNN structure following TSSL-BP~\cite{zhang2020temporal} and SLAYER~\cite{shrestha2018slayer}. Each input example of the original N-MNIST dataset is over 300,000 time steps. We reduced the time resolution to 100 time steps following the procedure described in~\cite{zhang2020temporal}. STBP~\cite{wu2019direct} applied spike accumulation instead, and reduced the time resolution to 300 time steps. To demonstrate the strength of the proposed NA method, we further compressed each input example by retaining only the beginning  $30\%$ portion of the input spike trains to train a spiking CNN based on the architecture specified in Table~\label{tab:N-MNIST} using the NA algorithm over a short time window of 30 time points. And yet, our NA algorithm outperforms all other methods based on the same spiking CNN architecture as reported in Table~\ref{tab:N-MNIST}. 

\begin{table}[h]
    \centering
    \vspace{-2mm}
    \caption{Performances of different SNN BP methods on N-MNIST.}
    \begin{tabular}{c|c|c}
         \hline
         Methods  & \#Time steps & Best accuracy \\
         \hline
         SLAYER & 300 & 99.22\% \\
         TSSL-BP & 100 & 99.40\% \\
         TSSL-BP   & 30 & 99.28\% \\
         This work & 30 & \textbf{99.35\%} \\
         \hline
         \multicolumn{3}{l}{Spiking CNN structure: 12C5-P2-64C5-P2}\\
    \end{tabular}
    \vspace{-2mm}
    \label{tab:N-MNIST}
\end{table}



\subsection{The Results on the CIFAR10 Dataset}
\begin{table}[b]
    \centering
    \vspace{-5mm}
    \caption{Performances of different SNN BP methods on  CIFAR10.}    
    \begin{tabular}{c|c|c|c}
         \hline
         Methods  & Structure & \#Time steps & Best accuracy \\
         \hline
         STBP & AlexNet & 12 & 85.24\% \\
         STBP & CifarNet & 12 & 90.53\% \\ 
         TSSL-BP & AlexNet & 5 & 89.22\% \\
        \textbf{This work} & AlexNet & 5 & \textbf{91.76\%} \\
         \hline
         \multicolumn{4}{l}{AlexNet structure: 96C3-256C3-P2-384C3-}\\
         \multicolumn{4}{l}{P2-384C3-256C3-1024-1024}\\
         \multicolumn{4}{l}{CifarNet structure: 128C3-256C3-P2-512C3-}\\
         \multicolumn{4}{l}{P2-1024C3-
512C3-1024-512}
    \end{tabular}
    \label{tab:CIFAR10}
\end{table}
The CIFAR10 dataset~\cite{krizhevsky2009learning} is  a challenging  image dataset for testing direct training of SNNs.
Table~\ref{tab:CIFAR10} compares the accuracy of SNNs based on the widely adopted Alexnet and CifarNet architectures  trained by three different BP methods.  Within a short time window of five time steps, the Alexnet model trained by our NA algorithm had the best accuracy, which even surpassed the performance of the larger CifarNet model trained by the STBP algorithm~\cite{wu2019direct} over a longer time window. 

\subsection{Time Complexity Analysis}

The time complexity of both STBP~\cite{wu2019direct} and TSSL-BP is $\mathcal{O}(N_t)$, where $N_t$ is the number of time steps. With neighborhood selection using SNS, NA perturbs the membrane potential of each spiking neuron $N_t$ times and computes
the resulting  neighbors by running Algorithm~\ref{algo:add}  $N_t$ times. Since a single run of Algorithm~\ref{algo:add} has a complexity of $\mathcal{O}(N_t)$, the overall complexity of NA is $\mathcal{O}(N_t^2)$.

\begin{figure}[h]
    \centering
    \includegraphics[width=0.45\textwidth]{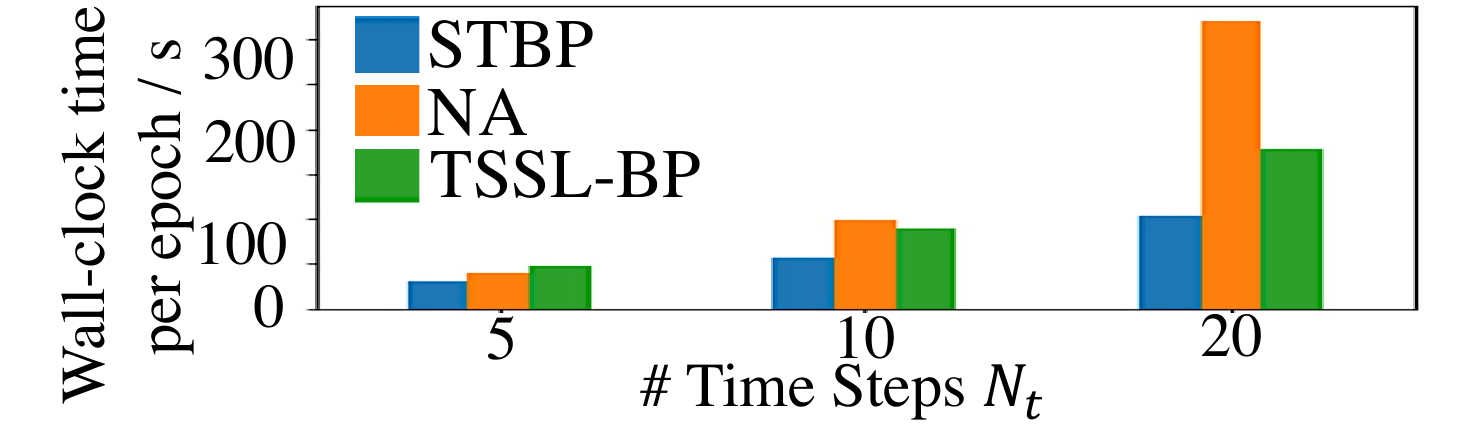}
    \vspace{-0.2cm}
    \caption{Wall-clock time comparison (MNIST dataset, single TITAN XP GPU, BatchSize=64)}
    \label{fig:time_complex}
    \vspace{-0.5cm}
\end{figure}

Figure \ref{fig:time_complex} reports the wall-clock times consumed by one training epoch of STBP, TSSL-BP, and NA, respectively, for training the spiking CNNs of Table~\ref{tab:MNIST}.  
While having a higher theoretical complexity, NA is faster than TSSL-BP~\cite{zhang2020temporal} when $N_t=5$, which may be due to differences in code optimization. 
The complexity of NA may be lowered by choosing fewer but more important membrane potential neighbors or limiting the range of temporal dependencies considered in Algorithm~\ref{algo:add}, which will be explored in our future work.

\section{Conclusion}
We presented a new SNNs direct training method, neighborhood aggregation (NA),  consisting of three key components: membrane potential neighborhood computation, a membrane potential distance measure, and a computationally-efficient backpropagated neighborhood aggregation pipeline. NA avoids  calculating the conventionally defined error gradient that does not always exist, and instead gears towards leveraging finite differences within a neighborhood of each neuron's membrane potential. This leads to the so-called aggregated gradient that is more meaningful for spiking neurons.
The superiority of the NA algorithm over the prior arts is demonstrated by benchmarking on commonly used datasets. This work provides a general finite-difference based direct SNN training framework, and  leaves great flexibility for developing improved methods of neighborhood selection and finite difference aggregation.


\bibliography{reference}

\begin{thebibliography}{32}
\providecommand{\natexlab}[1]{#1}
\providecommand{\url}[1]{\texttt{#1}}
\expandafter\ifx\csname urlstyle\endcsname\relax
  \providecommand{\doi}[1]{doi: #1}\else
  \providecommand{\doi}{doi: \begingroup \urlstyle{rm}\Url}\fi

\bibitem[Bellec et~al.(2018)Bellec, Salaj, Subramoney, Legenstein, and
  Maass]{bellec2018long}
Bellec, G., Salaj, D., Subramoney, A., Legenstein, R., and Maass, W.
\newblock Long short-term memory and learning-to-learn in networks of spiking
  neurons.
\newblock \emph{Advances in neural information processing systems}, 2018.

\bibitem[Berner et~al.(2019)Berner, Brockman, Chan, Cheung, D{\k{e}}biak,
  Dennison, Farhi, Fischer, Hashme, Hesse, et~al.]{berner2019dota}
Berner, C., Brockman, G., Chan, B., Cheung, V., D{\k{e}}biak, P., Dennison, C.,
  Farhi, D., Fischer, Q., Hashme, S., Hesse, C., et~al.
\newblock Dota 2 with large scale deep reinforcement learning.
\newblock \emph{arXiv preprint arXiv:1912.06680}, 2019.

\bibitem[Blouw et~al.(2019)Blouw, Choo, Hunsberger, and
  Eliasmith]{blouw2019benchmarking}
Blouw, P., Choo, X., Hunsberger, E., and Eliasmith, C.
\newblock Benchmarking keyword spotting efficiency on neuromorphic hardware.
\newblock In \emph{Proceedings of the 7th Annual Neuro-inspired Computational
  Elements Workshop}, pp.\  1--8, 2019.

\bibitem[Bohte et~al.(2002)Bohte, Kok, and La~Poutre]{bohte2002error}
Bohte, S.~M., Kok, J.~N., and La~Poutre, H.
\newblock Error-backpropagation in temporally encoded networks of spiking
  neurons.
\newblock \emph{Neurocomputing}, 48\penalty0 (1-4):\penalty0 17--37, 2002.

\bibitem[Brown et~al.(2020)Brown, Mann, Ryder, Subbiah, Kaplan, Dhariwal,
  Neelakantan, Shyam, Sastry, Askell, Agarwal, Herbert-Voss, Krueger, Henighan,
  Child, Ramesh, Ziegler, Wu, Winter, Hesse, Chen, Sigler, Litwin, Gray, Chess,
  Clark, Berner, McCandlish, Radford, Sutskever, and Amodei]{brown2020language}
Brown, T.~B., Mann, B., Ryder, N., Subbiah, M., Kaplan, J., Dhariwal, P.,
  Neelakantan, A., Shyam, P., Sastry, G., Askell, A., Agarwal, S.,
  Herbert-Voss, A., Krueger, G., Henighan, T., Child, R., Ramesh, A., Ziegler,
  D.~M., Wu, J., Winter, C., Hesse, C., Chen, M., Sigler, E., Litwin, M., Gray,
  S., Chess, B., Clark, J., Berner, C., McCandlish, S., Radford, A., Sutskever,
  I., and Amodei, D.
\newblock Language models are few-shot learners.
\newblock 2020.

\bibitem[Davies et~al.(2018)Davies, Srinivasa, Lin, Chinya, Cao, Choday, Dimou,
  Joshi, Imam, Jain, et~al.]{davies2018loihi}
Davies, M., Srinivasa, N., Lin, T.-H., Chinya, G., Cao, Y., Choday, S.~H.,
  Dimou, G., Joshi, P., Imam, N., Jain, S., et~al.
\newblock Loihi: A neuromorphic manycore processor with on-chip learning.
\newblock \emph{IEEE Micro}, 38\penalty0 (1):\penalty0 82--99, 2018.

\bibitem[Deng et~al.(2020)Deng, Wu, Hu, Liang, Ding, Li, Zhao, Li, and
  Xie]{deng2020rethinking}
Deng, L., Wu, Y., Hu, X., Liang, L., Ding, Y., Li, G., Zhao, G., Li, P., and
  Xie, Y.
\newblock Rethinking the performance comparison between snns and anns.
\newblock \emph{Neural Networks}, 121:\penalty0 294--307, 2020.

\bibitem[Furber et~al.(2014)Furber, Galluppi, Temple, and
  Plana]{furber2014spinnaker}
Furber, S.~B., Galluppi, F., Temple, S., and Plana, L.~A.
\newblock The spinnaker project.
\newblock \emph{Proceedings of the IEEE}, 102\penalty0 (5):\penalty0 652--665,
  2014.

\bibitem[Gerstner \& Kistler(2002)Gerstner and Kistler]{gerstner2002spiking}
Gerstner, W. and Kistler, W.~M.
\newblock \emph{Spiking neuron models: Single neurons, populations,
  plasticity}.
\newblock Cambridge university press, 2002.

\bibitem[He et~al.(2015)He, Zhang, Ren, and Sun]{he2015delving}
He, K., Zhang, X., Ren, S., and Sun, J.
\newblock Delving deep into rectifiers: Surpassing human-level performance on
  imagenet classification.
\newblock In \emph{Proceedings of the IEEE international conference on computer
  vision}, pp.\  1026--1034, 2015.

\bibitem[Huh \& Sejnowski(2017)Huh and Sejnowski]{huh2017gradient}
Huh, D. and Sejnowski, T.~J.
\newblock Gradient descent for spiking neural networks.
\newblock \emph{Advances in neural information processing systems}, 2017.

\bibitem[Jin et~al.(2018)Jin, Zhang, and Li]{jin2018hybrid}
Jin, Y., Zhang, W., and Li, P.
\newblock Hybrid macro/micro level backpropagation for training deep spiking
  neural networks.
\newblock \emph{Advances in neural information processing systems},
  31:\penalty0 7005--7015, 2018.

\bibitem[Kim et~al.(2020)Kim, Kim, and Kim]{kim2020unifying}
Kim, J., Kim, K., and Kim, J.-J.
\newblock Unifying activation-and timing-based learning rules for spiking
  neural networks.
\newblock \emph{Advances in Neural Information Processing Systems}, 2020.

\bibitem[Krizhevsky et~al.(2009)Krizhevsky, Hinton,
  et~al.]{krizhevsky2009learning}
Krizhevsky, A., Hinton, G., et~al.
\newblock Learning multiple layers of features from tiny images.
\newblock 2009.

\bibitem[Krizhevsky et~al.(2017)Krizhevsky, Sutskever, and
  Hinton]{krizhevsky2017imagenet}
Krizhevsky, A., Sutskever, I., and Hinton, G.~E.
\newblock Imagenet classification with deep convolutional neural networks.
\newblock \emph{Communications of the ACM}, 60\penalty0 (6):\penalty0 84--90,
  2017.

\bibitem[LeCun(1998)]{lecun1998mnist}
LeCun, Y.
\newblock The mnist database of handwritten digits.
\newblock \emph{http://yann. lecun. com/exdb/mnist/}, 1998.

\bibitem[Loshchilov \& Hutter(2017)Loshchilov and
  Hutter]{loshchilov2017decoupled}
Loshchilov, I. and Hutter, F.
\newblock Decoupled weight decay regularization.
\newblock \emph{arXiv preprint arXiv:1711.05101}, 2017.

\bibitem[Merolla et~al.(2014)Merolla, Arthur, Alvarez-Icaza, Cassidy, Sawada,
  Akopyan, Jackson, Imam, Guo, Nakamura, et~al.]{merolla2014million}
Merolla, P.~A., Arthur, J.~V., Alvarez-Icaza, R., Cassidy, A.~S., Sawada, J.,
  Akopyan, F., Jackson, B.~L., Imam, N., Guo, C., Nakamura, Y., et~al.
\newblock A million spiking-neuron integrated circuit with a scalable
  communication network and interface.
\newblock \emph{Science}, 345\penalty0 (6197):\penalty0 668--673, 2014.

\bibitem[Mostafa(2017)]{mostafa2017supervised}
Mostafa, H.
\newblock Supervised learning based on temporal coding in spiking neural
  networks.
\newblock \emph{IEEE transactions on neural networks and learning systems},
  29\penalty0 (7):\penalty0 3227--3235, 2017.

\bibitem[Niu et~al.(2020)Niu, Li, Li, Xiao, Sun, Deng, and Chen]{niu2020dual}
Niu, X., Li, B., Li, C., Xiao, R., Sun, H., Deng, H., and Chen, Z.
\newblock A dual heterogeneous graph attention network to improve long-tail
  performance for shop search in e-commerce.
\newblock In \emph{Proceedings of the 26th ACM SIGKDD International Conference
  on Knowledge Discovery \& Data Mining}, pp.\  3405--3415, 2020.

\bibitem[Orchard et~al.(2015)Orchard, Jayawant, Cohen, and
  Thakor]{orchard2015converting}
Orchard, G., Jayawant, A., Cohen, G.~K., and Thakor, N.
\newblock Converting static image datasets to spiking neuromorphic datasets
  using saccades.
\newblock \emph{Frontiers in neuroscience}, 9:\penalty0 437, 2015.

\bibitem[Rossum(2001)]{rossum2001novel}
Rossum, M.~v.
\newblock A novel spike distance.
\newblock \emph{Neural computation}, 13\penalty0 (4):\penalty0 751--763, 2001.

\bibitem[Shorten \& Khoshgoftaar(2019)Shorten and
  Khoshgoftaar]{shorten2019survey}
Shorten, C. and Khoshgoftaar, T.~M.
\newblock A survey on image data augmentation for deep learning.
\newblock \emph{Journal of Big Data}, 6\penalty0 (1):\penalty0 1--48, 2019.

\bibitem[Shrestha \& Orchard(2018)Shrestha and Orchard]{shrestha2018slayer}
Shrestha, S.~B. and Orchard, G.
\newblock Slayer: Spike layer error reassignment in time.
\newblock \emph{Advances in Neural Information Processing Systems}, 2018.

\bibitem[Victor \& Purpura(1997)Victor and Purpura]{spike_distances}
Victor, J.~D. and Purpura, K.~P.
\newblock Metric-space analysis of spike trains: theory, algorithms and
  application.
\newblock \emph{Network: Computation in Neural Systems}, 8\penalty0
  (2):\penalty0 127--164, 1997.

\bibitem[Wu et~al.(2018)Wu, Deng, Li, Zhu, and Shi]{wu2018spatio}
Wu, Y., Deng, L., Li, G., Zhu, J., and Shi, L.
\newblock Spatio-temporal backpropagation for training high-performance spiking
  neural networks.
\newblock \emph{Frontiers in neuroscience}, 12:\penalty0 331, 2018.

\bibitem[Wu et~al.(2019)Wu, Deng, Li, Zhu, Xie, and Shi]{wu2019direct}
Wu, Y., Deng, L., Li, G., Zhu, J., Xie, Y., and Shi, L.
\newblock Direct training for spiking neural networks: Faster, larger, better.
\newblock In \emph{Proceedings of the AAAI Conference on Artificial
  Intelligence}, volume~33, pp.\  1311--1318, 2019.

\bibitem[Yang(2020)]{yang2020temporal}
Yang, Y.
\newblock Temporal surrogate back-propagation for spiking neural networks.
\newblock \emph{arXiv preprint arXiv:2011.09964}, 2020.

\bibitem[Ye et~al.(2020)Ye, Chen, Zhang, Chen, Yuan, Liu, Chen, Liu, Qiu, Yu,
  et~al.]{ye2020towards}
Ye, D., Chen, G., Zhang, W., Chen, S., Yuan, B., Liu, B., Chen, J., Liu, Z.,
  Qiu, F., Yu, H., et~al.
\newblock Towards playing full moba games with deep reinforcement learning.
\newblock \emph{Advances in Neural Information Processing Systems}, 2020.

\bibitem[Zenke \& Ganguli(2018)Zenke and Ganguli]{zenke2018superspike}
Zenke, F. and Ganguli, S.
\newblock Superspike: Supervised learning in multilayer spiking neural
  networks.
\newblock \emph{Neural computation}, 30\penalty0 (6):\penalty0 1514--1541,
  2018.

\bibitem[Zhang \& Li(2019)Zhang and Li]{zhang2019spike}
Zhang, W. and Li, P.
\newblock Spike-train level backpropagation for training deep recurrent spiking
  neural networks.
\newblock \emph{Advances in neural information processing systems}, 2019.

\bibitem[Zhang \& Li(2020)Zhang and Li]{zhang2020temporal}
Zhang, W. and Li, P.
\newblock Temporal spike sequence learning via backpropagation for deep spiking
  neural networks.
\newblock \emph{Advances in Neural Information Processing Systems}, 33, 2020.

\end{thebibliography}

\clearpage
\newpage

\normalsize
\appendix

\section{Detailed Experimental Setups}

\subsection{Finite Difference Scaling and Clipping in Aggregated Gradient Computation}
Experimentally, it has been revealed that neurons' membrane potentials $\boldsymbol{u}$ mostly transfer to their closest neighbors during training, making the finite differences  involving farther neighbors less important and sometimes even misleading. Furthermore, the third step in the NA pipeline computes the first-order approximation to the loss change with respect to the PSC, which becomes less accurate when there exists a large distance between the neuron's present $\boldsymbol{a}$ and its neighbor's PSC $\boldsymbol{a}_p$.    Based on the two reasons above, instead of using the finite difference $f_{d(\boldsymbol{u},\boldsymbol{u}_p)}L$ defined in (\ref{eqn_finite_diff}), we employed the following scaled finite difference $\widetilde{f}_{d(\boldsymbol{u},\boldsymbol{u}_p)}L$ for aggregated gradient computation:   

\begin{equation}
\begin{aligned}
    \widetilde{f}_{d(\boldsymbol{u},\boldsymbol{u}_p)}L&= \boldsymbol{e}\cdot \left(\boldsymbol{a} _p-\boldsymbol{a} \right) \\
    &\cdot clip \left( \frac{1}{\mathrm{d_{MP}}(\boldsymbol{u}, \boldsymbol{u}_p)^3}, -b, b\right),
    \label{scaled_eqn_finite_diff}
\end{aligned}
\end{equation}


In order to avoid value explosions when $\mathrm{d_{MP}}(\boldsymbol{u}, \boldsymbol{u}_p)$ approaches zero, we clipped $\frac{1}{\mathrm{d_{MP}}(\boldsymbol{u}, \boldsymbol{u}_p)^3}$ within $[-b,b]$.
We recommend to set the hyperparameter $b\in[2,20]$ for stable performance, which was set to 10 in all our experiments.

\subsection{Training setups for different datasets} 
The proposed NA algorithm was run on a single Nvidia Titan Xp GPU to train SNNs based on three different datasets. The specific settings used are described below. 

\subsubsection{MNIST}
The MNIST dataset \cite{lecun1998mnist} contains 60,000 training images and 10,000 testing images.  We set the batch size to 64, the number of training epochs to 200, and the learning rate to 0.0005 for the adopted AdamW optimizer \cite{loshchilov2017decoupled}. The images were converted to  continuous-valued multi-channel  currents applied as the inputs to the SNN under training.    Moreover, data augmentations using  RandomCrop and RandomRotation were applied to improve performance \cite{shorten2019survey}. 

\subsubsection{N-MNIST}
The N-MNIST  \cite{orchard2015converting} is the neuromorphic version of the MNIST dataset  \cite{lecun1998mnist} and also has 60,000 training images and 10,000 testing images. 
We trained an SNN using the NA algorithm for 100 epochs  with a batch size of 50. The AdamW optimizer \cite{loshchilov2017decoupled} with a learning rate of 0.0005 was applied. No data augmentation was used. 

\subsubsection{CIFAR10}

The CIFAR10 dataset \cite{krizhevsky2009learning} contains 50,000 training images and 10,000 test images.
We trained our SNN using NA for 600 epochs with a batch size of 50 and a learning rate 0.0005 for the AdamW optimizer \cite{loshchilov2017decoupled}. 
The input image coding strategy used for the MNIST dataset was adopted. Moreover, data augmentations including RandomCrop, ColorJitter, and RandomHorizontalFlip \cite{shorten2019survey} were applied. The convolutional layers were initialized using the kaiming uniform initializer \cite{he2015delving}, and the linear layers were initialized using the kaiming normal initializer \cite{he2015delving}.

\bibliographystyle{icml2021}

\end{document}